\documentclass[fleqn,table]{wlscirep}

\usepackage{color}
\usepackage{multibib}
\usepackage{array}
\usepackage{amssymb}
\usepackage{afterpage}

\usepackage{tikz}
\usepackage{eso-pic}

\usepackage{amsmath}

\usepackage{floatrow}
\usepackage{enumitem}
\usepackage{multicol}

\usepackage[author={Manfred Eppe}]{pdfcomment}

\newcommand{\todo}[1]{}

\newcommand{\concept}[1]{\emph{#1}}

\usepackage{lipsum}

\usepackage{booktabs} 
\usepackage{siunitx} 
\usepackage{pgfplotstable} 

\usepackage[breakable, theorems, skins]{tcolorbox}
\tcbset{enhanced}

\usepackage{scrextend}
\changefontsizes[10pt]{9pt}
\usepackage{pgf}
\pgfdeclareradialshading{shadednodered}
{\pgfpoint{-0.2cm}{0.5cm}}%
{rgb(0cm)=(1,1,1);
rgb(0.9cm)=(0.9,0.3,0.1); rgb(1.4cm)=(0.7,0.2,0); rgb(4.4cm)=(0,0,0)}

\pgfdeclareradialshading{shadednodeorange}
{\pgfpoint{-0.2cm}{0.5cm}}%
{rgb(0cm)=(1,1,1);
rgb(0.9cm)=(0.95,0.5,0.2); rgb(1.4cm)=(0.7,0.35,0.1); rgb(4.4cm)=(0,0,0)}

\pgfdeclareradialshading{shadednodeyellow}
{\pgfpoint{-0.2cm}{0.5cm}}%
{rgb(0cm)=(1,1,1);
rgb(0.9cm)=(0.95,0.8,0.5); rgb(1.4cm)=(0.7,0.6,0.3); rgb(4.4cm)=(0,0,0)}

\pgfdeclareradialshading{shadednodewhite}
{\pgfpoint{-0.2cm}{0.5cm}}%
{rgb(0cm)=(1,1,1);rgb(4.4cm)=(1,1,1)}

\tikzset{
  LabelStyle/.style = { rectangle, rounded corners, 
                        draw,
                        fill = gray!50, 
                        font = \small},
TextStyle/.style = { font = \small},                        
  VertexStyle/.append style = { inner sep=5pt,
                                font = \small\color{black},
                                shape = rectangle, 
                                rounded corners,
                                minimum height = 1.5cm,
                                color=white,
                                shading = shadednodered,
                                align = center},
  EdgeStyle/.append style = {->, align = center} }
  
\usepackage{array}
\newcolumntype{L}[1]{>{\raggedright\let\newline\\\arraybackslash\hspace{0pt}}m{#1}}
\newcolumntype{C}[1]{>{\centering\let\newline\\\arraybackslash\hspace{0pt}}m{#1}}
\newcolumntype{R}[1]{>{\raggedleft\let\newline\\\arraybackslash\hspace{0pt}}m{#1}}
\usepackage{multirow}
\usepackage{colortbl}
\usepackage{makecell}

\definecolor{TableRowColor}{rgb}{0.9821,0.953,0.906}
\definecolor{TableHeadColor}{rgb}{0.9843,0.92157,0.82745}

\DeclareCaptionLabelSeparator{vertpipe}{ | }                
\usepackage{float}

\usepackage{titlesec}

\titleformat{\section}
  {\color{color1}\large\sffamily\bfseries}
  {}
  {0.0em}
  {#1}
  []

\titleformat{\subsection}
  {\sffamily\bfseries}
  {}
  {0.0em}
  {#1}
  []

\fancypagestyle{empty}{
  \lhead{This is an early preprint of the paper \textit{Intelligent problem-solving as integrated hierarchical reinforcement learning}, now published in Nature Machine Intelligence. Please cite and consult the new version at Nature Machine Intelligence \url{https://doi.org/10.1038/s42256-021-00433-9} or the latest preprint on arxiv \url{}, not this one.}
}

\title{Hierarchical principles of embodied reinforcement learning: A review}

\author[1,*]{Manfred Eppe}
\author[2,3]{Christian Gumbsch}
\author[1]{Matthias Kerzel}
\author[1]{Phuong D.H. Nguyen}
\author[2]{Martin V. Butz}
\author[1]{Stefan Wermter}
\affil[1]{Universität Hamburg, Germany}
\affil[2]{Universität Tübingen, Germany}
\affil[3]{Max Planck Institute for Intelligent Systems, Tübingen, Germany}
\affil[*]{eppe@informatik.uni-hamburg.de}

\begin{abstract}
Cognitive Psychology and related disciplines have identified several critical mechanisms that enable intelligent biological agents to learn to solve complex problems. 
There exists pressing evidence that the cognitive mechanisms that enable problem-solving skills in these species build on hierarchical mental representations. 
Among the most promising computational approaches to provide comparable learning-based problem-solving abilities for artificial agents and robots is hierarchical reinforcement learning. 
However, so far the existing computational approaches have not been able to equip artificial agents with problem-solving abilities that are comparable to intelligent animals, including human and non-human primates, crows, or octopuses.
Here, we first survey the literature in Cognitive Psychology, and related disciplines, and find that many important mental mechanisms involve compositional abstraction, curiosity, and forward models.
We then relate these insights with contemporary hierarchical reinforcement learning methods, and identify the key machine intelligence approaches that realise these mechanisms. 
As our main result, we show that all important cognitive mechanisms have been implemented independently in isolated computational architectures, and there is simply a lack of approaches that integrate them appropriately. 
We expect our results to guide the development of more sophisticated cognitively inspired hierarchical methods, so that future artificial agents achieve a problem-solving performance on the level of intelligent animals. 
\end{abstract}

\begin{document}
\twocolumn

\flushbottom
\maketitle

\thispagestyle{empty}

\begin{figure*}[h]
    \centering
    \includegraphics[width=.19\columnwidth,trim={10.93cm 5.52cm 10.93cm 3.52cm},clip]{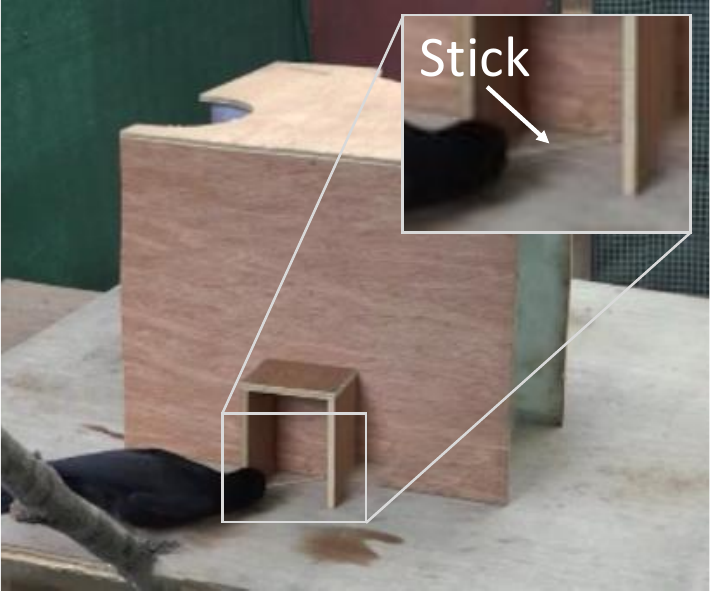}
    \hspace{-0.4cm}\begin{minipage}[b][0.3cm][t]{0.3cm}\textbf{a}\end{minipage} 
    \includegraphics[width=.19\columnwidth,trim={10.93cm 5.52cm 10.93cm 3.52cm},clip]{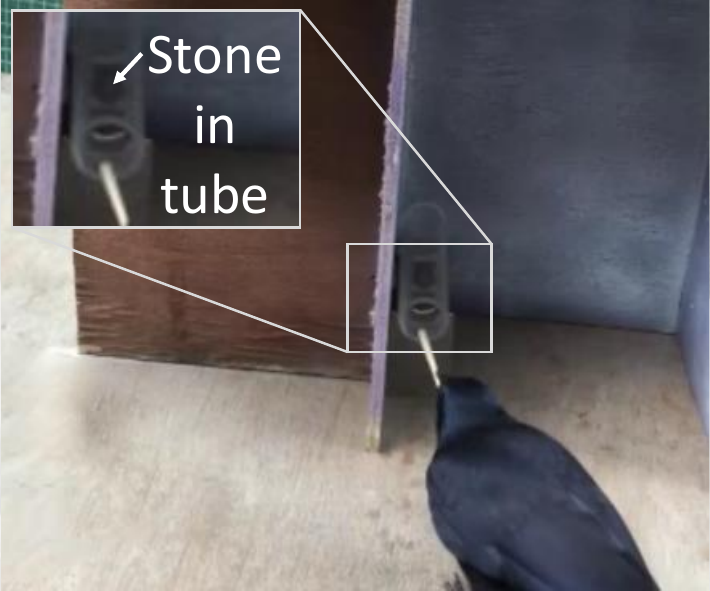}
    \hspace{-0.4cm}\begin{minipage}[b][0.3cm][t]{0.3cm}\textbf{b}\end{minipage} 
    \includegraphics[width=.19\columnwidth,trim={10.93cm 5.52cm 10.93cm 3.52cm},clip]{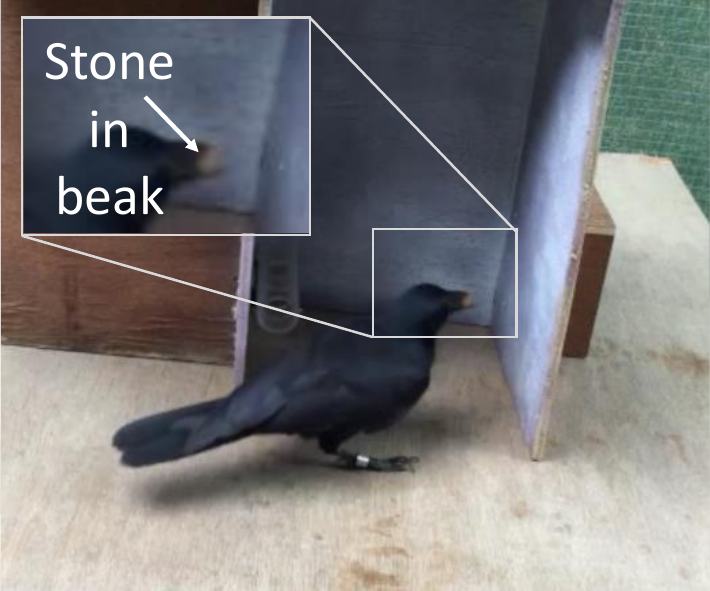}
    \hspace{-0.4cm}\begin{minipage}[b][0.3cm][t]{0.3cm}\textbf{c}\end{minipage} 
    \includegraphics[width=.19\columnwidth,trim={10.93cm 5.52cm 10.93cm 3.52cm},clip]{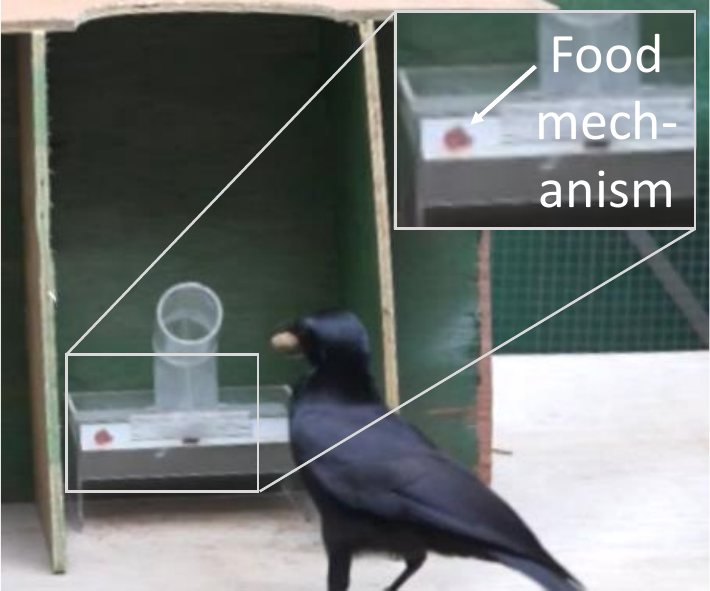}
    \hspace{-0.4cm}\begin{minipage}[b][0.3cm][t]{0.3cm}\textbf{d}\end{minipage} 
    \includegraphics[width=.19\columnwidth,trim={10.93cm 5.52cm 10.93cm 3.52cm},clip]{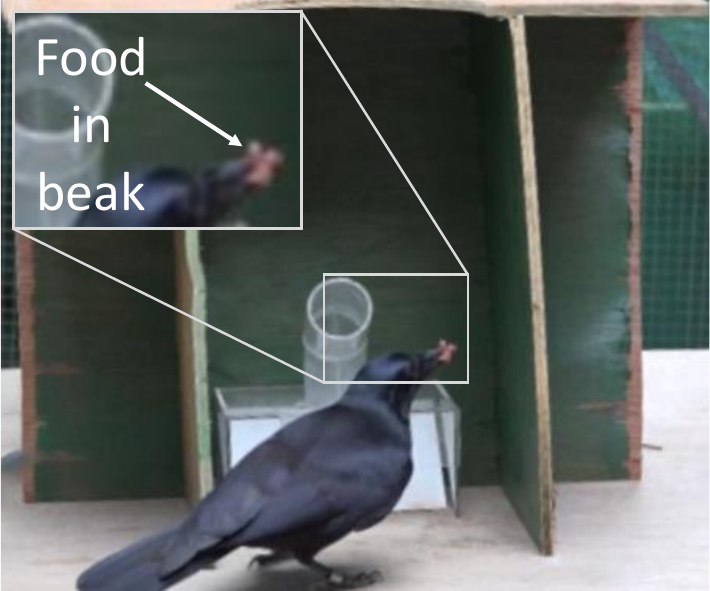}
    \hspace{-0.4cm}\begin{minipage}[b][0.3cm][t]{0.3cm}\textbf{e}\end{minipage} 
    \vspace{-10pt}
    \caption{A New Caledonian crow solves a food-access problem:\cite{Gruber2019_CrowMentalRep} \textbf{a,} First, the crow picks a stick. \textbf{b-c,} Then it uses the stick to pull a stone out of a tube. \textbf{d-e,} Finally, it uses the stone to activate a mechanism that releases food. The crow has never solved this problem setup  before. Nevertheless, after a brief inspection phase, it is able to solve the problem, suggesting that the crow develops memorized abstract mental representations to plan ahead.}.
    \label{fig:crow_probsolving}
\end{figure*}



Humans and several other higher level intelligent animal species have the ability to break down complex unknown problems into hierarchies of simpler previously learned sub-problems. This hierarchical approach allows them to solve previously unseen problems in a zero-shot manner, i.e., without any trial and error. 
For example, \autoref{fig:crow_probsolving} depicts how a New Caledonian crow solves a non-trivial food-access puzzle that consists of three causal steps: It first picks a stick, then uses the stick to access a stone, and then uses the stone to activate a mechanism that releases food \cite{Gruber2019_CrowMentalRep}. 
There exist numerous analogous experiments that attest similar capabilities to primates, octopuses, and, of course,  humans \cite{Butz2017_MindBeing,Perkins1992}.
This raises the question of how we can equip intelligent artificial agents and robots with similar problem-solving abilities.
To answer this question, the involved computational mechanisms and algorithmic implementation options need to be identified.

A very general computational framework for learning-based problem-solving is reinforcement learning (RL) \cite{Arulkumaran2017_DeepRL,Li2018_DeepRL_Overview,Sutton:2018}. 
Several studies suggest that RL is in many aspects biologically and cognitively plausible \cite{Neftci2019_RL_BioArtificial_NatureMI,Sutton:2018}. 
Existing RL-based methods are to some extent able to perform zero-shot problem-solving and transfer learning.
However, this is currently only possible for minor variations of the same or a similar task \cite{Eppe2019_planning_rl} or in simple synthetic domains, such as a 2D gridworld \cite{Oh2017_Zero-shotTaskGeneralizationDeepRL,Sohn2018_Zero-ShotHRL}.
A continuous-space problem-solving behaviour that is comparable with the crow's behaviour of \autoref{fig:crow_probsolving} has not yet been realised with any artificial system, based on RL or other approaches.

Research in human and animal cognition strongly suggests that problem-solving and learning is hierarchical \cite{Botvinick2009_HierarchicalCognitionRL,Butz2017_MindBeing,Tomov2020}. 
We hypothesise that one reason why current machine learning systems fail is that the existing approaches underestimate the power of learning hierarchical abstractions. 
At the time of writing this article, we performed a comprehensive meta-search over RL review articles from 2015 to 2020 using the Microsoft Academic search engine. Our meta-survey has yielded the following results:
The most cited review article on RL since 2015 \cite{Arulkumaran2017_DeepRL} dedicates 1/6th of a page out of 16 pages to hierarchical approaches.
In the second most cited article \cite{Garcia2015_safeRLSurvey}, hierarchical RL is not considered at all, i.e, the word stem `hierarchic' does not appear anywhere in the text. The third most cited review article \cite{Li2018_DeepRL_Overview} dedicates 2/3rd of a page out of 85 total pages to hierarchical RL.
From 37 RL reviews and surveys published since 2015, only two contain the word stem ``hierarchic'' in their abstract. 
The second edition of the popular RL book by \citet{Sutton:2018} only mentions hierarchies in the twenty-year-old options framework \cite{Sutton1999_options} on 2 pages in the final book chapter and briefly discusses automatic abstraction a few lines later in that chapter. 
It appears that researchers are struggling with automatically learning RL-suitable hierarchical structures. 


%

In this article, we address this gap by illuminating potential reasons for this struggle and by providing pointers to solutions.
We show that most computational hierarchical RL approaches are model-free, whereas the results from our survey of biological cognitive models suggest that suitable predictive forward models are needed.
Furthermore, we exhibit that existing hierarchical RL methods hardly address state abstraction and compositional representations. 
However, we also show that there exist exceptions where isolated relevant cognitive mechanisms including forward models and compositional abstraction have already been implemented, but not in combination with other important mechanisms.
As a result, we conclude that the AI community already has most of the tools for building more intelligent agents at hand;
but it currently lacks holistic efforts to integrate them appropriately.
%

%



\begin{figure*}[h!]
    \centering
        \includegraphics[width=\textwidth]{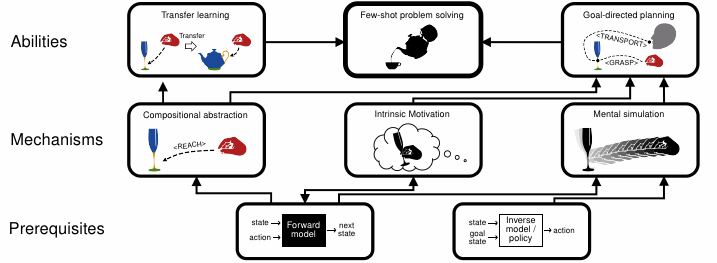}
    \caption{Prerequisites, mechanisms, and features of biological problem-solving agents. 
    \emph{Forward-} and \emph{inverse models} are prerequisites for higher-order mechanisms and abilities.
    \emph{compositional abstractions} of forward models allow the agent to decompose a problem into reusable substructures, e.g., by learning which part of an object is easily graspable, such as the stem of a glass.
    \emph{Intrinsic motivation} applied with the help of forward models can drive an agent to interact with its environment in an information seeking, epistemic way, e.g., by eliciting interactions with the glass to learn more about its properties.
    \emph{Mental simulations} enable the exploration of hypothetical state-action sequences, e.g., by imagining how the hand position will evolve while reaching for the glass.
    Given these mechanisms, an agent can flexibly achieve desired goal states via hierarchical, compositional \emph{goal-directed planning}, e.g., planning to drink from the glass by first suitably grasping and then appropriately transporting it.
    Compositional abstractions enable the agent to perform \emph{transfer learning}, e.g., by executing suitable grasps onto other objects with stem-like parts, such as the shown teapot. 
    Taken together, these abilities enable \emph{few-shot problem-solving}.}
    \label{fig:cog_prereq}
\end{figure*}

\section{Neurocognitive foundations}
\label{sec:cognitive_background}
The problem-solving abilities of a biological agent depend on cognitive key mechanisms that include abstraction, intrinsic motivation, and mental simulation.
\autoref{fig:cog_prereq} shows how these mechanisms depend on forward and inverse models as neuro-functional prerequisites.
Together, cognitive abilities that we deem crucial for higher level intelligence become enabled, including few-shot problem solving and transfer learning. 
In the following, we start with a characterization of these crucial cognitive abilities and then exhibit some of the key mechanisms and prerequisites needed to enable them.

\subsection{Crucial cognitive abilities}
By cognitive abilities for problem-solving we refer to phenomenologically observable traits and properties of biological agents.
As the most remarkable ability, we consider few-shot problem-solving.

\smallskip
\noindent
\textbf{Few-shot problem-solving} is the ability to solve unknown problems with few ($\lessapprox 10$) trials.\footnote{Zero-shot problem-solving is a special case of few-shot problem-solving where no additional training at all is required to solve a new problem.}
%
For simple problems, few-shot problem-solving is trivial. For example, catching a ball is usually a purely reactive behaviour that builds on a direct neural mechanism to map the observation of the flying ball to appropriate motor commands. Such problems are solvable with current computational RL methods\cite{Li2018_DeepRL_Overview}. 
However, there are also more difficult problems that require, e.g., using tools in new contexts. 
For example, consider the aforementioned problem-solving example of the  crow\cite{Gruber2019_CrowMentalRep} (see \autoref{fig:crow_probsolving}). The crow knows how to use a stick as a tool without any further training, because it has previously used a stick for related problems.
In our conceptual framework (see \autoref{fig:cog_prereq}) we identify two abilities that are most central to perform such problem-solving, namely transfer learning and planning. 

\smallskip
\noindent
\textbf{Transfer learning}
 allows biological agents to perform few-shot pro\-blem-solving by transferring the solution to a previously solved task to novel, previously unseen tasks\cite{Perkins1992}. This significantly reduces and sometimes eliminates the number of trials to solve a problem. \citet{Perkins1992} propose a distinction between near and far transfer. Near transfer refers to the transfer of skills between domains or situations that are rather similar, e.g. learning to catch a ball and then catching other objects. In contrast, far transfer requires the transfer of more abstract solutions between different situations typically via abstract analogies.  
Such analogies are determined by mappings between problems from different domains. 
For example, an object transport problem with a robotic hand that involves [\concept{grasp(object)}, \concept{moveArm(goal)}, \concept{release(object)}] is analogous to a file movement problem on a computer that involves \concept{[cutInFolder(file)}, \concept{ navigate(targetFolder)}, \concept{pasteInFolder(file)}], with the obvious mappings between the action types and arguments. 

Cognitive theories consider analogical reasoning as a critical cornerstone of human cognition. For example, conceptual metaphor theory\cite{Feldman2009_neural_ECG} understands metaphors as mappings between analogous concepts in different domains.
In addition to evidence from linguistics, there is also significant evidence from education theory and research on mechanical reasoning that analogical reasoning improves problem-solving: Education theory suggests that human transfer learning improves when explicitly trained to identify analogies between different problems\cite{Klauer1989_AnalogyTransfer}. 
The Psychology of mechanical inference suggests that analogical knowledge about the dynamic properties of mechanical objects is often transferred between domains\cite{Hegarty2004_MechReasMentalSim}. 
For example, knowledge about what happens if somebody jumps into a pool can be transferred to other mechanical problems that involve over-flooding water containers\cite{Hegarty2004_MechReasMentalSim}.

\smallskip
\noindent
\textbf{Goal-directed planning.}
The behaviour of biological agents is traditionally divided into two categories: Stimulus-driven, habituated behaviour and goal-directed, planned behaviour\cite{Dolan2013, Friston2016}.
Initially, stimulus-response theories dominated the field, suggesting that most behaviour was learned by forming associations that were previously reinforced \cite{Thorndike}.
Edward Tolman was one of the main critiques of stimulus-response theories\cite{Dolan2013}. He showed that rats can find rewards in a maze faster when they have visited the maze before, even if their previous visit had not been rewarded \cite{Tolman1930}.
The results suggest that the rats form a representation, or cognitive map, of the maze, which enables them to plan or mentally simulate their behaviour once a reward was detected.
Habituation of behaviour, which is comparable to model-free policy optimization in reinforcement learning\cite{Botvinick2014,Dayan2009}, enables an agent to learn one particular skill in a controlled environment.
However, complex problem-solving in a new or changing environment requires some form of prospective planning\cite{Butz2017_MindBeing, Dolan2013, Hoffmann:2003}.
Over the last decades, more research has focused on understanding the mechanisms and representations involved in goal-directed planning.
It is now generally agreed upon that, during planning, humans predict the effect of their actions, compare them to the desired effects, and, if required, refine their course of actions\cite{Hoffmann:2003,Kunde:2007}. 
This is deemed to be a hierarchical process where the effects of actions on different levels of abstraction are considered\cite{Botvinick2014, Tomov2020}. 
Thereby, humans tend to first plan high-level actions before considering actions at a finer granularity\cite{Wiener2003}.
Hierarchical planning can dramatically decrease the computational cost involved in planning \cite{Botvinick2014, Dolan2013}.
Additionally, hierarchical abstractions enable automatizing originally planned behaviour, thus further alleviating computational cost \cite{Dayan2009}.


\subsection{Cognitive mechanisms for transfer learning and planning}
Transfer learning and planning rely on shared and partially overlapping mechanisms and principles. Here, we focus on three mechanisms and principles that we identify as most critical, namely sensorimotor abstraction, intrinsic motivation, and mental simulation.

\subsection{Sensorimotor abstraction.}
\begin{figure}
\begin{minipage}{.32\textwidth}
\centering
\includegraphics[width=\textwidth]{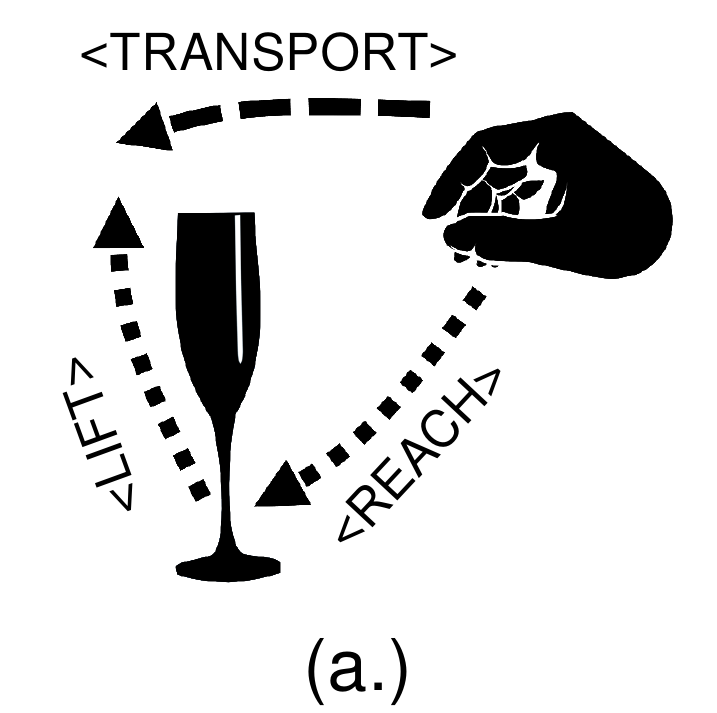}
\end{minipage}
\hfill
\begin{minipage}{.32\textwidth}
\centering
\includegraphics[width=\textwidth]{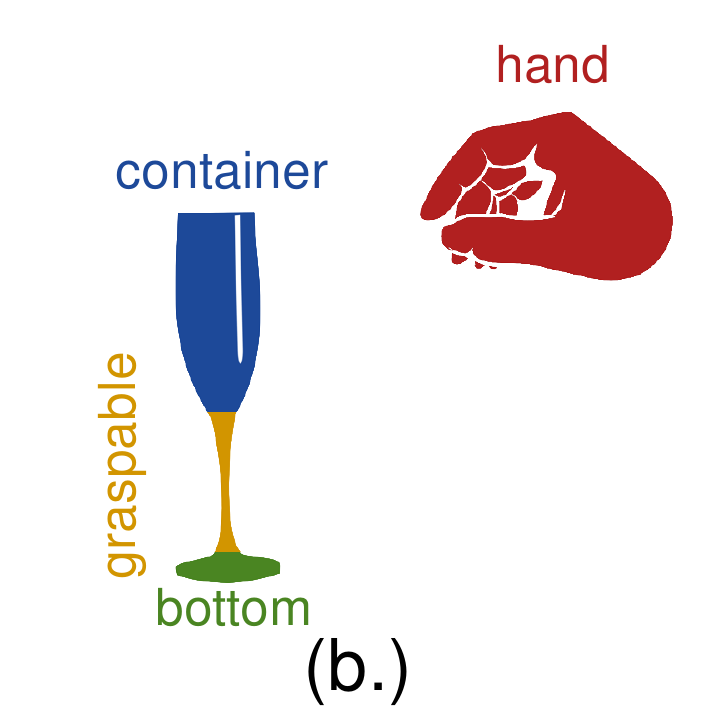}
\end{minipage}
\hfill
\begin{minipage}{.32\textwidth}
\centering
\includegraphics[width=\textwidth]{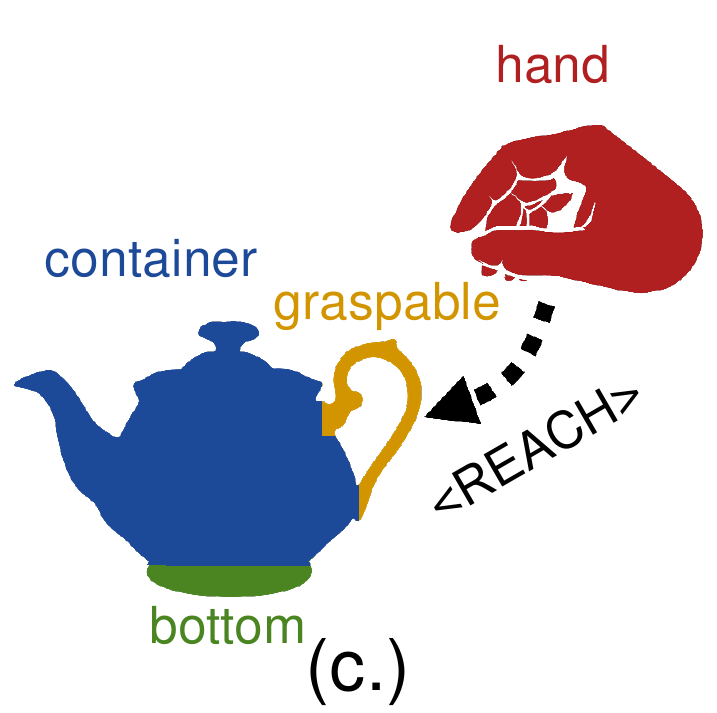}
\end{minipage}
\caption{Compositional action and state abstractions. \emph{Action abstractions} describe a sequence of more primitive actions (a.). \emph{State abstractions} encode certain properties of the state space (b). Their \emph{compositionality} enables their general application, by instantiating abstract definitions (the graspable target of reaching), with a specific object (teapot) (c.). \label{fig:cog_comp}}
\end{figure}

According to the widely accepted embodiment theory, abstract mental concepts are inferred from our bodily-grounded sensorimotor environmental interaction experiences \cite{Barsalou2008_GroundedCognition,Butz2017_MindBeing,Butz2016,Pulvermuller:2010}. 
Cognitive theories often distinguish between \emph{action abstraction} and \emph{state abstraction}.
Action abstractions refer to a temporally extended sequence of primitive actions, that are often referred to as options\cite{Sutton1999_options} or movement primitives\cite{Flash2005_AnimalMotorPrimitives, Schaal_2006}.
For example, \emph{transporting an object} can be seen as an action abstraction since it is composed of a sequence of more primitive actions, such as \textit{reaching}, \textit{lifting}, and \textit{transporting} (see \autoref{fig:cog_comp}).
The elementary actions that compose an action abstraction are typically rather loosely defined through their effects.
For example, we can tell that a robot arm we have never seen before is grasping an object, even though the anatomy, the unfolding control algorithms, and the involved motor commands may largely vary from our human ones.
State abstractions refer to encoding certain parts of the environment while abstracting away irrelevant details.
A simple example of a state abstraction is a code that specifies whether a certain type of object is present in a scene. 
State abstractions can be hierarchically structured:
A single object can be subdivided into its parts to form a partonomy, i.e., a hierarchical organization of its parts and their spatial relationship (see \autoref{fig:cog_comp}) \cite{Minsky1974, Zacks2001_EventStructure}.
Additionally, objects can be organized taxonomically, reflecting a hierarchical relationship that focuses on particular conceptual aspects \cite{Minsky1974, Zacks2001_EventStructure}.
For example, an affordance-oriented \cite{Gibson:1979} taxonomy of beverage containers could classify \textit{wine glasses} and \textit{beer mugs} both as \textit{drinking vessels}, while a \textit{teapot} would be classified as a \textit{pot} instead.
Meanwhile, \textit{beer mugs}, \textit{wine glasses}, and \textit{teapots} are all \textit{graspable containers}.


Conceptual abstractions and cross-references to other conceptual abstractions within imply a representational key property: \textbf{compositionality}.
Formal compositionality principles state that an expression or representation is compositional if it consists of subexpressions, and if there exist rules that determine how to combine the subexpressions \cite{Szabo:2020}. 
The Language of Thought theory\cite{Fodor:2001} transfers the compositionality principle from language to abstract mental representations, claiming that thought must also be compositional. 
For example, humans are easily able to imagine previously unseen situations and objects by composing their known constituents, such as the \emph{golden pavements} and \emph{ruby walls} imagined by Scottish philosopher David Humes in \emph{A Treatise of Human Nature}\cite{Frankland2020_SearchLoT}.

On top of that, embodied world knowledge further constrains the formal semantics-based options to combined sub-expressions.
In the remainder, we will refer to this type of compositionality by \textbf{common sense compositionality}, that is, compositional rules that are grounded in and flexibly inferred by our intuitive knowledge about the world, including other agents (humans, animals, robots, machines, etc.).  
This common sense compositionality essentially enforces that our minds attempt to form plausible environmental scenarios and events that unfold within given conceptual information about a particular situation \cite{Butz2016,Gaerdenfors:2014,Lakoff1999}. 
Consider the processes that unfold when understanding a sentence like ``He sneezed the napkin off the table''.
Most people have probably never heard or used this sentence before, but everyone with sufficient world and English grammar knowledge can correctly interpret the sentence, imagining a scene wherein the described event unfolds \cite{Butz2016}. 
Common sense compositionality makes abstractions applicable in meaningful manners, by constraining the filling in of an abstract component (e.g., the  target of a grasp) towards applicable objects (e.g., a graspable object such as a teapot), as depicted in \autoref{fig:cog_comp}.
Note the awkwardness -- and indeed the attempt of the brain to resolve it -- when disobeying common sense compositionality, such as when stating that `The dax fits into the air' or `grasping a cloud' \cite{Butz2016}.

Common sense compositionality indeed may be considered a hallmark and cornerstone of human creativity, as has been (indirectly) suggested by numerous influential cognitivists \cite{Barsalou2008_GroundedCognition,Fillmore1985,Lake2017,Lakoff1999,Sugita:2011,Turner2014,Werning:2012}.
In the neurosciences, recent neuroimaging techniques indeed revealed that the compositionality of mental representations can actually be observed in neural codes\cite{Frankland2020_SearchLoT}.
For example, \citet{Haynes2015_CompositionalNeuroRep} and \citet{Reverberi2012_CompositionalFmrIRules} show that the neural codes in the ventrolateral prefrontal cortex representing compound behavioural rules can be decomposed into neural codes that encode their constituent rules. 
Similar results have been obtained for neural codes that represent food compositions\cite{Barron2013_CombinedChoicesConcepts}.

From an algorithmic perspective, common sense compositionality yields benefits for analogical reasoning and planning. 
It simplifies the identification of analogies because it enables compositional morphisms between representations. 
This advantage is well-known in cognitive theories of creativity, e.g., concept blending\cite{Eppe2018_AIJ_Blending,Turner2014}, where the search for analogies is a computational bottleneck.
Similarly, compositional state- and action representations for goal-directed planning lead to a lower computational complexity, because action and object types can be flexibly combined as long as critical properties match.
Along these lines, action grammars have been proposed, which systematically form an effect-oriented taxonomy of tool-object interactions \cite{Worgotter:2013}. 
As a result, common sense compositionality enables the utilisation of objects as tools in innovative manners---when, for example, utilising a stone as a hammer \cite{Butz2017_MindBeing,Lakoff1999}. 
Moreover, it enables drawing analogies across domains---when , for example, talking about ``holding onto a thought'' for later utilisation.

Common sense compositionality thus seems to be key for truly intelligent artificial systems.
We propose that apart from the addition of suitable inductive learning biases, such as event-segmentation tendencies \cite{Butz2016,Butz2017_MindBeing,Butz:2020,Shin:2020tsi,Zacks2007_EST}, 
suitably targeted intrinsic motivation and the ability to play out mental simulations are of critical importance.

\smallskip
\noindent
\textbf{Intrinsic motivation} affects goal-directed reasoning and planning because it sets intrinsically motivated goals that an agent may aim to achieve. 
The most basic behaviour of biological agents purely strives for satisfying homeostatic needs, such as hunger or thirst. However, advanced behaviour, such as exploration and play, seems to be partially decoupled from the primary biological needs of an animal.
From a cognitive development perspective, the term \emph{intrinsic motivation} was coined to describe the ``inherent tendency to seek out novelty, [...] to explore and to learn''\cite{RyanDeci2000}. 
Here, \emph{intrinsic} is used in opposition to extrinsic values that are directly associated with the satisfaction of needs or the avoidance of undesired outcomes \cite{Friston2015}.
Intrinsic motivations induce general exploratory behaviour, curiosity, and playfulness.
Simple exploratory behaviour can even be found in worms, insects, and crustaceans \cite{Pisula2008} and may be elicited by rather simple tendencies to wander around while being satiated or an inborn tendency to evade overcrowded areas.   
Curiosity refers to an epistemic drive that causes information gain-oriented exploratory behaviour\cite{Berlyne1966, Loewenstein1994, Oudeyer2007}.
Combined with developing common sense compositionality, this drive can elicit hypothesis testing behaviour even by means of analogy making. 
The closely related playfulness is only exhibited in certain intelligent, mostly mammalian or avian species, such as dogs and corvids\cite{Pisula2008}, where scenarios and events within are played out in a hypothetical manner and novel interactions are tried-out in a skill-improvement-oriented manner. 



\smallskip

\noindent
\textbf{Mental simulation}, meanwhile,
enables biological agents to reflect on and to anticipate the dynamics of its environment on various representational levels in the past, present, and future, and even fully hypothetically.
Actual motor-imagery has been shown to enable athletes to improve the execution of challenging body movements (e.g. a backflip), significantly reducing the number of actually executed trials\cite{Jeannerod1995_MentalImageryMotor}.
On the reasoning side, consider human engineers who rely on  mental simulation when developing mechanical systems, such as pulleys or gears.
The simulation involves visual imaginations, but also includes non-visual properties like friction and force \cite{Hegarty2004_MechReasMentalSim}. 
Mental simulation also takes place on higher conceptual, compositional, and causal reasoning levels. For example, \citet{Kahnemann1982_MentalSim}, \citet{Wells1989_MentalSimCausality} and later \citet{Taylor1998_MentalSim} report how mental simulation improves the planning of future behaviour on a causal conceptual level, such as when planning a trip. 

\smallskip

\subsection{Forward-inverse models as functional prerequisites}

Sensorimotor abstraction, intrinsic motivation, and mental simulation are complex mechanisms that demand a suitable neuro-functional basis. 
We propose that this essential basis is constituted by forward and inverse models \cite{Wolpert1998_ForwardInverse}. 
Forward models predict how an action affects the world, while inverse models determine the actions that need to be executed to achieve a goal or to obtain a reward.
Note that inverse models may be implicitly inferred from the available forward models, but more compact inverse models, which are closely related to habits and motion primitives, certainly foster the proficiency of particular behavioural skills further \cite{Barsalou2008_GroundedCognition,Butz2017_MindBeing,Dezfouli:2014,Schaal_2006}. 

\smallskip
\noindent
\textbf{Forward-inverse models for mental simulation.}
To perform mental simulation, an agent needs a forward model to simulate how the dynamics of the environment behave, possibly conditioned on the agent's actions.
However, even when equipped with a well-predicting forward model, the consideration of all possible actions and action sequences quickly becomes computationally intractable. 
Hence, action-selection requires more direct mappings from state and desired inner mental state (including an intention or a goal) to potential actions. 
This is accomplished by inverse models. 
In RL, inverse models are represented as behaviour policies.
Selecting actions can happen in a reflex-like manner on the motor level but it can also serve as a heuristic to generate candidate actions for accomplishing a current goal. 
For example, when designing a new mechanical system, engineers have an intuition about how to put certain mechanical parts together, depending on the goal function they want to achieve with the system. Based on their intuition, they mentally simulate and verify the functioning of the mechanical system\cite{Hegarty2004_MechReasMentalSim}. 

\smallskip

\noindent
\textbf{Forward and inverse models for intrinsic motivation.}
Forward and inverse models are also needed to trigger behaviour that is driven by intrinsic motivation. For example, prediction error minimisation has been demonstrated to be a useful additional, intrinsic reward signal to model curiosity, inversely triggering behaviour and behavioural routines that are believed to cause particular environmental changes that are, in turn, believed to possibly result in knowledge gain \cite{Kaplan:2004, Schmidhuber2010, Schmidhuber:1991, Oudeyer2007}.


Along similar lines, Friston et al.~ propose that intrinsically motivated behaviour emerges when applying active inference\cite{Friston2011, Friston2015}.
Active inference describes the process of inferring actions to minimize expected free energy, which includes approximations of anticipated surprise\cite{Friston2010, Friston2016}. 
Free energy is decomposed into various sources of uncertainty.
One part is uncertainty about future states or outcomes given a certain sequence of actions \cite{Friston2015}. 
The agent strives to reduce this uncertainty, and with it overall expected free energy, by activating the actions that cause the uncertainty.
Hence, active inference can lead to epistemic exploration, where an agent acts out different behaviour to learn about its consequences\cite{Friston2015, Oudeyer2018_CompuCuriosityLearning}.
Here, the forward model is required to predict future states given an action, and also to estimate the uncertainty of the prediction.

\smallskip
\noindent
\textbf{Forward and inverse models for abstraction and grounding.}
Over the last decade, various theories including predictive coding\cite{Huang2011}, the Free Energy Principle\cite{Friston2010}, and the Bayesian Brain hypothesis\cite{Knill2004}, have viewed the brain as a generative machine, which constantly attempts to match incoming sensory signals with its probabilistic predictions\cite{Clark2013}.
Within these frameworks, prediction takes place on multiple processing hierarchies that interact bidirectionally:
Top-down information per processing stage provides additional information to predict sensory signals, while bottom-up error information is used to correct the top-down predictions \cite{Clark2016_SurfUncertainty}.
Meanwhile, future states are predicted in a probabilistic manner\cite{Clark2013}.
On lower processing levels, actual incoming sensorimotor signals are predicted, while on higher levels more abstract, conceptual and compositional predictive encodings emerge\cite{Butz2016}.
In this way, rather complex state abstractions, such as the aforementioned \concept{container}-concept, can form. 
As a result, high-level predictions, e.g., of an object being within a container, enable predictions on lower levels.
For example, consider the  prediction of sensory information about how an object's position will change over time while it is occluded.

Event Segmentation Theory (EST)\cite{Zacks2007_EST}  makes the role of forward predictions even more explicit:
EST is concerned with why humans seem to perceive the continuous sensory activity in terms of discrete, conceptual events.
The theory suggests that these segmentations mirror the internal representation of the unfolding experience. 
Internal representations of events, or \emph{event models}, are sensorimotor abstractions that encode entities with certain properties and their functional relations in a spatiotemporal framework\cite{Radvansky2011_EventPerception}.
According to EST, these event models guide the perception by providing additional information that can be used for forward predictions\cite{Zacks2007_EST}.

While observing an event, a specific subset of event models is active until a transient prediction error is registered, resulting in an exchange of the currently active event models  to a new subset that may be better suited for predicting the currently ongoing dynamics\cite{Kuperberg:2020tsi, Radvansky2011_EventPerception, Shin:2020tsi, Zacks2007_EST}.
EST-inspired computational models demonstrate that such transient forward prediction errors can indeed be used to signal event transitions in video streams or self-explored sensorimotor data in an online fashion
\cite{Franklin:2020, Gumbsch2019, Humaidan:2020}.


\section{Computational realizations}
\label{sec:hrlFinal}

\begin{figure}
    \centering
    \includegraphics[width=\textwidth]{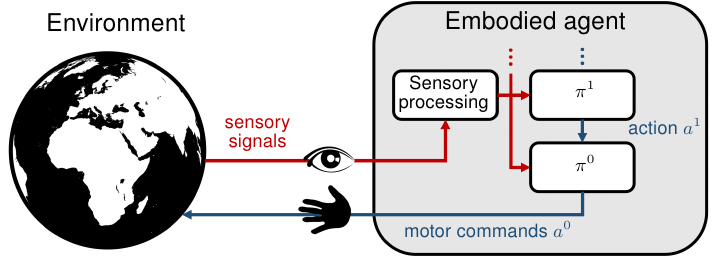} 
    \caption{A general hierarchical problem-solving architecture with a separate policy $\pi^i$ for each layer of abstraction. 
    Most existing computational HRL approaches focus on two layers, where the high-level action $a^1$ is either a behavioural primitive (an option or sub-policy), or a subgoal.
    Only the low-level actions $a^0$ are motor commands that affect the environment directly. }
    \label{fig:hrl}
\end{figure}

The abilities, mechanisms and prerequisites of computational hierarchical reinforcement learning systems (cf.~\autoref{fig:hrl}) are less sophisticated and integrated than those of biological agents. However, there are promising novel developments to potentially overcome the existing limitations. To identify the potential of the existing computational approaches, we provide an overview of the current state of the art on hierarchical reinforcement learning in \autoref{tab:hrl_properties}.
%


\begin{table*}[h!]
{\sffamily 
  \centering
  \captionsetup{singlelinecheck = false, justification = raggedright}
\caption{Abilities, mechanisms and prerequisites of recent hierarchical reinforcement learning  approaches.}
\vspace{-10pt}
  \scriptsize
\rowcolors{2}{TableRowColor}{TableRowColor}
  \begin{tabular}{R{.14\textwidth}  C{.05\textwidth}  C{.06\textwidth}  C{.08\textwidth}  C{.055\textwidth}  C{.055\textwidth}  C{.08\textwidth}  C{.08\textwidth}  C{.08\textwidth}  C{.08\textwidth} }
\arrayrulecolor{black}            
  \rowcolor{TableRowColor}
            & \multicolumn{3}{ c  }{\scshape \large Abilities} & \multicolumn{4}{ c  }{\scshape \large Mechanisms} & \multicolumn{2}{ c }{\scshape \large Prerequisites} \\
\rowcolor{TableRowColor}
  {\scshape \large Approach} & Few-shot problem-solving & Transfer learning --------------- \begin{minipage}{.06\textwidth} N: near \\ F: far \end{minipage} \newline & Goal-directed planning ------------------ \begin{minipage}{.08\textwidth} H: high-level \\ L: low-level \end{minipage} \newline &  \multicolumn{2}{ c }
   {
  \begin{minipage}{.13\textwidth} {\centering \vspace{5pt} Sensorimotor abstraction \\ --------------------------- \\} R: representational \\ C: compositional \\ \hspace{2pt}  ( actions )~~~~~~~~~ ( states )\end{minipage} 
   } & 
   Intrinsic motivation ------------------- \begin{minipage}{.1\textwidth}C: curiosity \\D: diversity \\SG: subgoals\\[0pt]  \end{minipage} &
   Mental simulation ------------------- \begin{minipage}{.08\textwidth} H: high-level \\ L: low-level \end{minipage} \newline & 
    \newline Forward model -------------------- \begin{minipage}{.08\textwidth} H: high-level \\ L: low-level \end{minipage} \newline & 
   Inverse model / policy ------------------- \begin{minipage}{.08\textwidth} H: high-level \\ L: low-level \end{minipage} \newline \\\hline
\arrayrulecolor{white}
\citet{Akrour2018_RL_StateAbstraction} & - & - & - & R,C & R & - & - & - & L,H\\\hline
\citet{Arulkumaran2016_Classifying_Options_for_Deep_RL} & - & - & - & R & - & - & - & - & L,H\\\hline
\citet{Bacon2017_OptionCritic} & - & - & - & R & - & - & - & - & L,H\\\hline
\citet{Beyret2019_DotToDot} & - & - & - & C & - & - & - & - & L,H\\\hline
\citet{Blaes2019_CWYC} & - & - & H & R & - & C, SG & H & H,L & L\\\hline
\citet{Chuck2020_HyPe} & - & - & - & R & - & C & - & - & L,H\\\hline
\citet{Dietterich2000_StateAbstraction_MAXQ} & - & - & - & R & R,C & SG & - & - & L,H\\\hline
\citet{Eppe2019_planning_rl} & - & - & H & R,C & R,C & SG & H & H & L\\\hline
\citet{Eysenbach2019_DiversityFunction} & - & N & - & R & - & D & - & - & L,H\\\hline
\citet{Frans2018_MetaLearningHierarchies} & (X) & F & - & R & - & - & - & - & L,H\\\hline
\citet{Ghazanfari2020_AssociationRules} & - & - & - & R & R & - & - & - & L,H\\\hline
\citet{Ghosh2019_ActionableRep} & - & - & - & R & R & SG & - & - & L,H\\\hline
\citet{Haarnoja2018_LatentSpacePoliciesHRL} & - & - & - & R & - & SG & - & - & L,H\\\hline
\citet{Han2020_hierarchicalSelfOrga} & X & N & - & R & - & - & - & - & L,H\\\hline
\citet{Heess2016_transferModulControl} & X & F & - & R & - & - & - & - & L,H\\\hline
\citet{Hejna2020_MorphologicalTransfer} & X & F & - & C & - & - & - & - & L,H\\\hline
\citet{Jiang2019_HRL_Language_Abstraction}  & X & N & - & R,C & - & - & - & - & L,H\\\hline
\citet{Kulkarni2016_HDQN} & - & - & - & R & R,C & SG & - & - & L,H\\\hline
\citet{Levy2019_Hierarchical} & - & - & - & C & - & SG & - & - & L,H\\\hline
\citet{Li2017_efficient_learning} & - & - & H & R,C & - & - & H,L & H,L & H\\\hline
\citet{Li2020_SubPolicy} & X & N & - & R & - & - & - & - & L,H\\\hline
\citet{Lyu2019_SDRL} & - & - & H & R & R,C & SG & H & H & L\\\hline
\citet{Ma2020_MultiAgentHRL} & - & - & H & R & R & - & H & H & L\\\hline
\citet{Machado2017_PVF_Laplace_OptionDiscovery} & - & - & - & R & R & D & - & - & L,H\\\hline
\citet{Nachum2018_HIRO} & - & - & - & C & - & SG & - & - & L,H\\\hline
\citet{Oh2017_Zero-shotTaskGeneralizationDeepRL} & X & N & - & R,C & R & SG & - & - & L,H\\\hline
\citet{Qiao_2020_HRL-Driving} & - & - & - & R & - & SG & - & - & L,H\\\hline
\citet{Qureshi2020_CompAgnosticPol}	& -	& F& 	-& 	R,C	&-& 	SG,D& 	-& 	-& 	L,H \\\hline
\citet{Rafati2019_model-free_rep_learning} & - & - & - & C & - & SG & - & - & L,H\\\hline
\citet{Rasmussen2017_NeuralHRL} & - & N & - & R,(C) & R & SG & - & - & L,H\\\hline
\citet{Riedmiller2018_SAC-X} & - & - & - & R & - & SG & - & - & L,H\\\hline
\citet{Roeder2020_CHAC} & - & - & - & C & - & C,SG & - & - & L,H\\\hline
\citet{Saxe2017_HRL_Multitask_LMDP} & - & - & - & R,C & - & SG & - & - & L,H\\\hline
\citet{Schaul2013_Better_Generalization_with_Forecasts} & - & - & - & R & R,C & - & - & - & L,H\\\hline
\citet{Shankar2020_MotorPrimitives} & - & - & - & R & - & D & - & - & L,H\\\hline
\citet{Sharma2020_DADS} & X & N & H & R & - & D & H & H & L\\\hline
\citet{Sohn2018_Zero-ShotHRL} & X & N & - & R,C & R & SG & - & - & L,H\\\hline
\citet{Tessler2017_LifelongHRLMinecraft} & (X) & N & - & R,C & - & - & - & - & L,H\\\hline
\citet{Vezhnevets2017_Feudal} & - & - & - & R & R & - & - & - & L,H\\\hline
\citet{Vezhnevets2020_OPRE} & - & - & - & R,C & R,C & -  & - & - & L,H\\\hline
\citet{Wu2019_ModelPrimitives}	& -	& N	& -	& R	& -	& -	& L	& L	& L,H \\\hline
\citet{Wulfmeier2020_HierachicalCompositionalPolicies} & - & N & - & R & - & - & - & - & L,H\\\hline
\citet{Yamamoto2018_RL_Planning} & - & - & H & R,C & - & SG & H & H & L\\\hline
\citet{Yang2018_PEORL} & - & - & H & R & - & SG & H & H & L\\\hline
\citet{Yang2018_HierarchicalControl} & - & - & - & R & - & - & - & - & L,H\\\hline
\citet{Zhang2020_AdjancentSubgoals} & - & - & - & C & - & SG & - & - & L,H\\\hline
  \end{tabular}
 \normalsize
    \label{tab:hrl_properties}
}
\end{table*}



\subsection{Transfer learning and planning for few-shot abilities}
\label{sec:hrl:abilities}
Our survey of the neurocognitive foundations indicates that two foundational cognitive abilities for few-shot problem-solving are transfer learning and planning. 

\smallskip
\noindent
\textbf{Transfer learning} denotes the re-use of previously learned skills in new contexts and domains, where \emph{near} transfer learning denotes transfer between similar contexts and domains between source and target tasks, while \emph{far} transfer considers stronger dissimilarities \cite{Perkins1992}.
A significant fraction of the existing near transfer approaches build on learning re-usable low-level skills, which are referred to as behavioural primitives, options, or sub-policies.
For example, \citet{Li2020_SubPolicy} and \citet{Frans2018_MetaLearningHierarchies} present sub-policy-based hierarchical extensions to Proximal Policy Optimization (PPO)\cite{Schulman2017_PPO}. 
Their approaches enable the transfer of sub-policies to novel tasks within the same domain. 
\citet{Heess2016_transferModulControl} use Generalized Advantage Estimation (GAE)\cite{Schulman2015_GAE} to implement similar transferable sub-policies. However, the authors also consider transfer learning towards different types of tasks. For example, they transfer behaviour from a locomotion task to a more complex soccer task. 
\citet{Eysenbach2019_DiversityFunction} and \citet{Sharma2020_DADS} introduce diversity functions to learn diverse re-usable sub-policies that are independent of external rewards.

\citet{Tessler2017_LifelongHRLMinecraft} focus on the transfer of skills in lifelong learning, and \citet{Wu2019_ModelPrimitives} propose a model-based approach, where only the forward models are transferred to new tasks, but not the policies. 
\citet{Vezhnevets2016_STRAW} build on the automatic discovery of transferable macro-actions (plans) to solve problems in discrete 2D-environments, while  \citet{Jiang2019_HRL_Language_Abstraction} use natural language action representations to perform near transfer learning.
\citet{Qureshi2020_CompAgnosticPol} build on re-usable low-level policies for goal-conditioned hierarchical reinforcement learning. 

Research has not only considered transfer learning between different tasks but also between different robot and agent morphologies\cite{Devin2017_Transfer_RL,Frans2018_MetaLearningHierarchies,Hejna2020_MorphologicalTransfer}. 
We classify these approaches as \emph{far} transfer because the entire sensorimotor apparatus changes, which places the agent in a dissimilar context.
Furthermore, the methods that feature such cross-morphological transfer also perform far transfer learning across different application domains. For example, \citet{Frans2018_MetaLearningHierarchies} transfer navigation skills acquired in a discrete-space grid maze environment to a continuous-space ant maze environment. 

\smallskip
\noindent
\textbf{Planning} is a highly overloaded term that has different meanings in different  sub-disciplines of AI. In this paper, we refer to goal-directed planning in the sense of classical AI, as an abductive search for a sequence of actions that will lead to a specific goal by simulating the actions with an internal model of the domain dynamics. 
Planning enables one-shot problem-solving because the searching does not involve the physical execution of actions. The agent only executes the actions if the mental simulation verifies that the actions are successful.
In this sense, planning differs from model-based reinforcement learning, which usually refers to training a policy by simulating actions using a predictive forward model\cite{Sutton1990_Dyna}.

Hierarchical planning is a well-known paradigm in classical AI\cite{Nau2003_SHOP2}, but approaches that integrate planning with hierarchical reinforcement learning are rare. 
Some approaches integrate action planning with reinforcement learning by using an action planner for high-level decision-making and a reinforcement learner for low-level motor control\cite{Eppe2019_planning_rl,Lyu2019_SDRL,Ma2020_MultiAgentHRL,Sharma2020_DADS,Yamamoto2018_RL_Planning,Yang2018_PEORL}. 
These approaches underpin that planning is especially useful for high-level inference in discrete state-action spaces. 


\subsection{Mechanisms behind transfer learning and planning}
Our summary of the cognitive principles behind transfer learning and planning reveals three important mechanisms that are critical for the learning and problem-solving capabilities of biological agents, namely compositional sensorimotor abstraction, intrinsic motivation, and mental simulation.

\begin{figure}
    \centering
    \includegraphics[width=.99\linewidth,trim={20pt 0pt 100pt 40pt},clip]{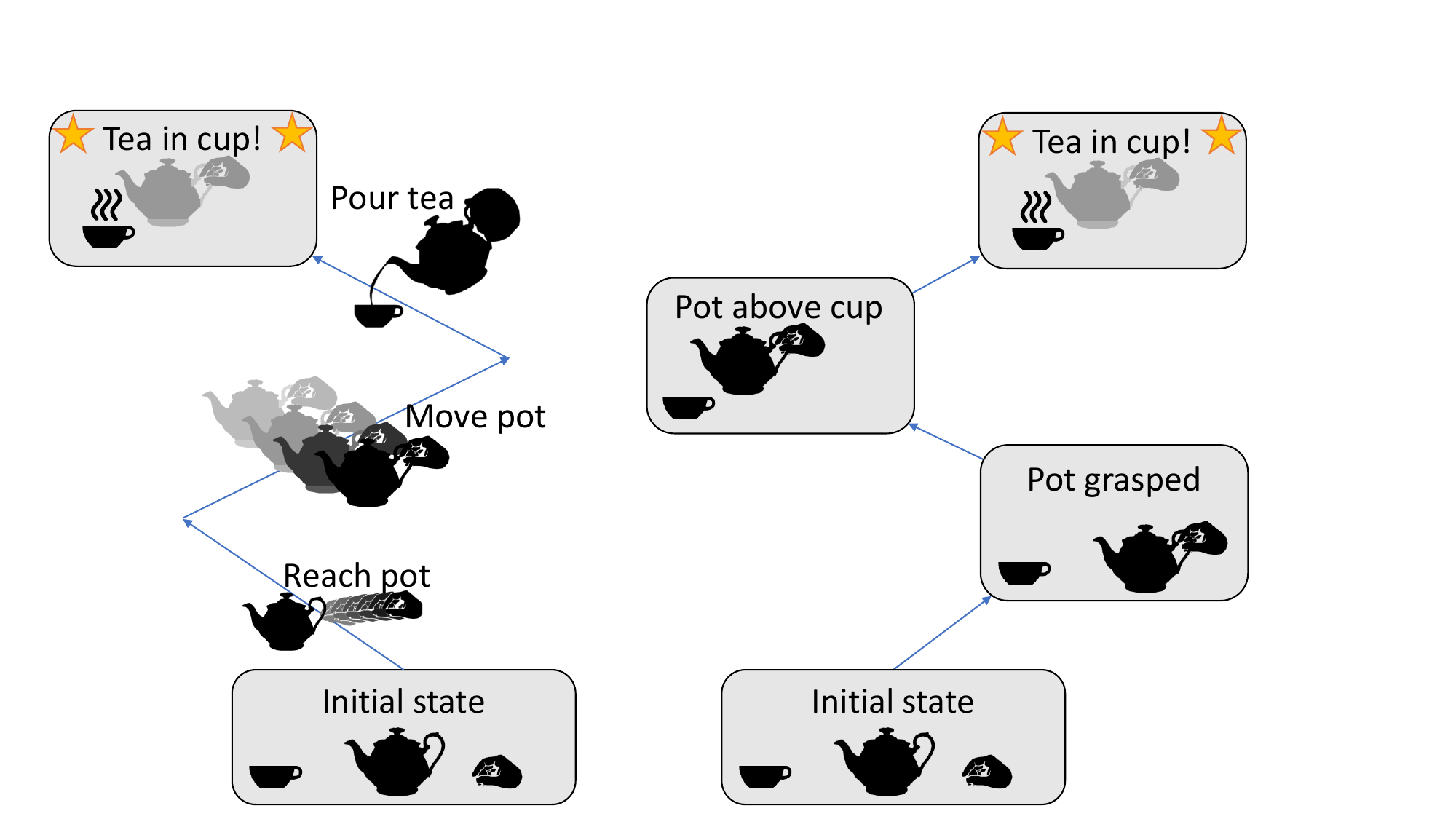}
    {\sf \footnotesize ~~~~~~~ (a.)    behaviour primitives  ~~~~~~~~~~~~ (b.) Subgoals ~~~~~~~~~~~~ }
    \caption{Action abstraction through behaviour primitives (a) vs subgoals (b). With behaviour primitives, the agent determines the path to the final goal by selecting a sequence of high-level actions, but without specifying explicitly to which intermediate state each action leads. With subgoals, the agent determines the intermediate states as subgoals that lead to the final goal, but without specifying the actions to achieve the subgoals.}
    \label{fig:options_behaviour_primitives}
\end{figure}

\smallskip
\noindent
\textbf{Compositional sensorimotor abstraction and grounding.}
The temporal abstraction of actions is, by definition, an inherent property of hierarchical reinforcement learning as it allows to break down complex problems into a temporal hierarchical structure of simpler problems. 
Another dimension of abstraction is representational abstraction. 
Hierarchical reinforcement learning approaches also involve representational action abstraction. 
One can distinguish the existing approaches into two different types of representational action abstraction.
The probably most influential method for action abstraction builds on behaviour primitives\cite{Bacon2017_OptionCritic,Dietterich2000_StateAbstraction_MAXQ,Frans2018_MetaLearningHierarchies,Kulkarni2016_HDQN,Li2020_SubPolicy,Li2017_efficient_learning,Ma2020_MultiAgentHRL,Sharma2020_DADS,Machado2017_PVF_Laplace_OptionDiscovery,Qiao_2020_HRL-Driving,Shankar2020_MotorPrimitives,Tessler2017_LifelongHRLMinecraft,Vezhnevets2020_OPRE}, including options\cite{Sutton1999_options}, sub-policies, or atomic high-level skills.
Such behaviour primitives are represented in an abstract representational space, e.g. in a discrete finite space of action labels or indices, abstracting away from the low-level space of motor commands (see \autoref{fig:options_behaviour_primitives}, a.). 
Another more recent type of producing high-level action representations are subgoals in the low-level state-space \cite{Eppe2019_planning_rl,Ghazanfari2020_AssociationRules,Levy2019_Hierarchical,Nachum2018_HIRO,Rafati2019_model-free_rep_learning}$^,$ ~ \cite{Hejna2020_MorphologicalTransfer,Roeder2020_CHAC,Zhang2020_AdjancentSubgoals} (see \autoref{fig:options_behaviour_primitives}, b.).
Subgoals are abstract actions defined in the low-level state space, and the agent achieves the final goal by following a sequence of subgoals.
There exist also methods that encode high-level actions as continuous vector representations. For example, \citet{Han2020_hierarchicalSelfOrga} use a multi-timescale RNN, where the high-level actions are encoded by the connections between the RNN layers and others use latent vector representations to encode high-level behaviour\cite{Isele2016_FeatureZeroShotTransfer,Shankar2020_MotorPrimitives}. 

Overall, in hierarchical reinforcement learning literature, there exists a strong implicit focus on action abstraction.
However significant cognitive evidence demonstrates that representational \emph{state abstraction} is at least as important to realize efficient problem-solving\cite{Butz2017_MindBeing,Lesort2018_state-rep-learning-control}. 
Yet, compared to action abstraction, there is considerably less research on state abstraction in hierarchical reinforcement learning. 
Cognitive state abstractions range from the mere preprocessing of sensor data, e.g. in the primary visual cortex, to the problem-driven grounding of signals in abstract compositional concept representations. Counterparts for some of these methods can also be found in computational architectures. 
For example, most current reinforcement learning approaches that process visual input use convolutional neural networks to preprocess the visual sensor data\cite{Jiang2019_HRL_Language_Abstraction,Lample2018_PlayingFPSGames,Oh2017_Zero-shotTaskGeneralizationDeepRL,Sohn2018_Zero-ShotHRL,Vezhnevets2017_Feudal,Wulfmeier2020_HierachicalCompositionalPolicies,Yang2018_HierarchicalControl}. 
A problem with simple preprocessing is that it does not appreciate that different levels of inference require different levels of abstraction: For example, to transport an object from one location to another with a gripper, only the low-level motor control layer of a hierarchical approach needs to know the precise shape and weight of the object. A high-level planning or inference layer only needs to know abstract Boolean facts, such as whether the object to transport is initially within the gripper's reach or not, and whether the object is light enough to be carried. 

Therefore, we consider only those approaches as representational abstraction methods that involve layer-wise abstraction. 
Layer-wise state abstraction has been tackled, but most existing hierarchical reinforcement learning approaches perform the abstraction using manually defined abstraction functions\cite{Eppe2019_planning_rl,Kulkarni2016_HDQN,Ma2020_MultiAgentHRL,Toussaint2018}.
There exist only a few exceptions where state abstractions are derived automatically in hierarchical reinforcement learning architectures\cite{Ghosh2019_ActionableRep,Vezhnevets2017_Feudal,Vezhnevets2020_OPRE}, e.g. through  clustering\cite{Akrour2018_RL_StateAbstraction,Ghazanfari2020_AssociationRules}, with feature representations\cite{Schaul2013_Better_Generalization_with_Forecasts} or by factorisation \cite{Vezhnevets2020_OPRE}. Interestingly, non-hierarchical model-based reinforcement learning offers promising alternative prediction-based methods for state abstraction \cite{Hafner2020_Dreamer,Pathak2019_ICML_SelfSuperExploreDisa}, which show parallels to cognitive prediction-based abstraction theories. 
However, these have not yet been applied in a hierarchical architecture. 

As implied from the cognitive science side, a key-property of representations of states and actions is \emph{compositionality}. 
For instance, a symbolic compositional action representation \concept{grasp(glass)} allows for modulating the action \emph{grasp} with the object \emph{glass}.
Compositionality is not limited to symbolic expressions, but also applicable to distributed numerical expressions, such as vector representations. For example, a vector $v_1$ is a compositional representation if it is composed of other vectors, e.g., $v_2$ and $v_3$, and if there is a vector operation $\circ$ that implements interpretation rules, e.g., $v_1 = v_2 \circ v_3$. 

There is significant cognitive evidence that compositionality improves transfer learning\cite{Colas2020_LanguageGroundingRL}. 
This evidence is computationally plausible when considering that transfer learning relies on exploiting analogies between problems. The analogies between two or more  problems in goal-conditioned reinforcement learning are defined by a multidimensional mapping between the initial state space, the goal space, and the action space of these problems. 
For example, given $n_a=4$ action types (e.g. \concept{``grasp'', ``push'', ``move'', ``release''}) and $n_o=4$ object types (e.g. \concept{``glass'', ``cup'', ``tea pot'', ``spoon''}), a non-compositional action-representations requires one distinct symbol for each action-object combination, resulting in $n_o \cdot n_a/2$ possible action mappings. In contrast, an analogy mapping with compositional actions would require to search over possible mappings between action types and, separately, over mappings between objects. Hence, the size of the search space is $n_o/2 + n_a/2$. 
Evidence that a lower number of possible analogy mappings improves transferability is also provided by other cognitively inspired computational methods, such as concept blending\cite{Eppe2018_AIJ_Blending}.

The work by \citet{Jiang2019_HRL_Language_Abstraction} provides further empirical evidence that compositional representations improve learning transfer. 
The authors use natural language as an inherently compositional representation to describe high-level actions in hierarchical reinforcement learning. 
In an ablation study, they find that the compositional natural language representation of actions improves transfer learning performance compared to non-compositional representations.

Few other researchers use compositional representations in hierarchical reinforcement learning. \citet{Saxe2017_HRL_Multitask_LMDP} compose high-level actions from concurrent linearly solvable Markov decision processes (LMDPs) to guarantee optimal compositionality. Their method successfully executes compositional policies that it has never executed before, effectively performing zero-shot problem-solving.
Zero-shot problem-solving has also been demonstrated by other related research that features compositional action representations, but that does not draw an explicit link between compositionality and zero-shot problem-solving\cite{Isele2016_FeatureZeroShotTransfer,Oh2017_Zero-shotTaskGeneralizationDeepRL,Sohn2018_Zero-ShotHRL}.
Few symbolic compositional state representations exist, but these rely on manually defined abstraction functions\cite{Eppe2019_planning_rl,Kulkarni2016_HDQN,Lyu2019_SDRL,Toussaint2018} or they are very general and feature compositionality only as an optional property\cite{Rasmussen2017_NeuralHRL}.
There also exist hierarchical reinforcement learning approaches where sub-symbolic compositional representations are learned\cite{Devin2017_Transfer_RL,Qureshi2020_CompAgnosticPol,Schaul2013_Better_Generalization_with_Forecasts,Vezhnevets2020_OPRE}. 
The compositionality of these representations is rather implicit and, with one exception \cite{Qureshi2020_CompAgnosticPol}, has not been investigated in the context of transfer learning. The exceptional approach by \citet{Qureshi2020_CompAgnosticPol} considers composable low-level policies and shows that compositionality significantly improves transfer between similar environments. 

\smallskip
\noindent
\textbf{Intrinsic motivation}
is a useful method to stabilise reinforcement learning by supplementing sparse external rewards. 
It is also commonly used to incentivize exploration.  
Reinforcement learning typically models intrinsic motivation through intrinsic rewards. The most common method of hierarchical reinforcement learning to generate intrinsic rewards is by providing intrinsic rewards when subgoals or subtasks are achieved\cite{Blaes2019_CWYC, Cabi2017_intentionalunintentional,Eppe2019_planning_rl,Haarnoja2018_LatentSpacePoliciesHRL,Jaderberg2017_unreal,Kulkarni2016_HDQN,Lyu2019_SDRL,Oh2017_Zero-shotTaskGeneralizationDeepRL,Qureshi2020_CompAgnosticPol,Rasmussen2017_NeuralHRL,Riedmiller2018_SAC-X,Roeder2020_CHAC,Saxe2017_HRL_Multitask_LMDP,Sohn2018_Zero-ShotHRL,Yamamoto2018_RL_Planning,Yang2018_PEORL}. 
Other approaches provide intrinsic motivation to identify a collection of behavioural primitives with a high diversity\cite{Blaes2019_CWYC, Eysenbach2019_DiversityFunction,Machado2017_PVF_Laplace_OptionDiscovery} and predictability\cite{Sharma2020_DADS}. 
This includes also the identification of primitives that are suited for re-composing high-level tasks\cite{Shankar2020_MotorPrimitives}.

Another prominent intrinsic reward model that is commonly used in non-hierarchical reinforcement learning is based on surprise and curiosity\cite{Oudeyer2007,Pathak2017_forward_model_intrinsic,Schillaci2020_IntrinsicMotivation,Schmidhuber2010}. 
In these approaches, surprise is usually modelled as a function of the prediction error of a forward model, and curiosity is realised by providing intrinsic rewards if the agent is surprised.
However, there is only little work on modelling surprise and curiosity in hierarchical reinforcement learning. Only \citet{Blaes2019_CWYC}, \citet{Colas2019_CURIOUS}, and \citet{Roeder2020_CHAC} use surprise in a hierarchical setting, showing that hierarchical curiosity leads to a significant improvement of the learning performance.
These approaches train a high-level layer to determine explorative subgoals in hierarchical reinforcement learning.
An alternative method to model curiosity is to perform hypothesis-testing for option discovery\cite{Chuck2020_HyPe}.

\smallskip
\noindent
\textbf{Mental simulation} is a mechanism that allows an agent to anticipate the effects of its own and other actions. Therefore, it is a core mechanism to equip an intelligent agent with the ability to plan ahead. Cognitive theories about mental simulation involve predictive coding\cite{Huang2011} and mental motor imagery\cite{Jeannerod1995_MentalImageryMotor}, while computational approaches to mental simulation involve model-based reinforcement learning\cite{Sutton1990_Dyna}, action planning\cite{Nau2003_SHOP2}, or a combination of both \cite{Eppe2019_planning_rl,Hafez2020_dual-system}. 
However, even though there is significant cognitive evidence that mental simulation happens on multiple representational layers\cite{Butz2017_MindBeing,Frankland2020_SearchLoT}, there is a lack of approaches that use hierarchical mental simulation in hierarchical reinforcement learning. 
Only a few approaches that integrate planning with reinforcement learning build on mental simulation\cite{Eppe2019_planning_rl,Lyu2019_SDRL,Ma2020_MultiAgentHRL,Sharma2020_DADS,Yamamoto2018_RL_Planning,Yang2018_PEORL}, though the mental simulation mechanisms of these models are only implemented on the high-level planning layer. 
An exception is the work by \citet{Wu2019_ModelPrimitives}, who use mental simulation on the low-level layer to determine the sub-policies to be executed. 
Another exception is presented by \citet{Li2017_efficient_learning} who perform mental simulation for planning on multiple task layers. 

There exist several non-hierarchical model-based reinforcement learning methods\cite{Ha2018_WorldModels,Hafner2020_Dreamer,Sutton1990_Dyna} that function akin to mental motor imagery: In these approaches, the policy is trained by mentally simulating action execution instead of executing the actions in the real world.
However, even though mental simulation is deemed to be a hierarchical process, we are not aware of any approach that performs hierarchical model-based reinforcement learning in the classical sense, where the policy is trained on the developing forward model.

\subsection{Prerequisites for sensorimotor abstraction, intrinsic motivation and mental simulation}
Reinforcement learning builds on policies that select actions based on the current observation and a goal or a reward function. Policies can be modelled directly, derived from value functions or combined with value functions.
In all cases, a policy is an inverse model that predicts the actions to be executed in the current state to maximise reward or to achieve a goal state.
In contrast, a forward model predicts a future world state based on the current state and a course of actions. 
Both kinds of models are critical prerequisites for the mechanisms that enable transfer learning, planning, and ultimately few-shot problem-solving.
Our review shows that the vast majority of hierarchical reinforcement learning methods use inverse models for both the high-level and low-level layers. 
However, some approaches exist that use an inverse model only for the low-level layer\cite{Eppe2019_planning_rl,Lyu2019_SDRL,Ma2020_MultiAgentHRL,Sharma2020_DADS,Yamamoto2018_RL_Planning,Yang2018_PEORL}. 
These frameworks use a planning mechanism, driven by a forward model, to perform the high-level decision-making.
Our review demonstrates that a forward model is required for several additional mechanisms that are necessary or at least highly beneficial for transfer learning, planning, and few-shot problem-solving. 
Some non-hierarchical approaches use a forward model to perform sensorimotor abstraction\cite{Hafner2020_Dreamer,Pathak2017_forward_model_intrinsic}. They achieve this with a self-supervised process where forward predictions are learned in a latent abstract space. 
However, we are not aware of any hierarchical method that exploits this mechanism. 

A forward model is also required to model curiosity as an intrinsic motivation mechanism. Cognitive science suggests that curiosity is one of the most important drives of human development, and it has been demonstrated to alleviate the sparse rewards problem in reinforcement learning\cite{Pathak2017_forward_model_intrinsic,Roeder2020_CHAC}. 
Curiosity is commonly modelled as a mechanism that rewards an agent for being surprised; surprise is formally a function of the error between the predicted dynamics of a system and the actual dynamics\cite{Friston2011}. To perform these predictions, a forward model is required.
However, with few exceptions\cite{Colas2019_CURIOUS,Roeder2020_CHAC}, there is a lack of approaches that use a hierarchical forward model for generating hierarchical curiosity. 


\section{Results}
\label{sec:results}
\autoref{tab:hrl_properties} shows the results of our review on computational hierarchical reinforcement learning approaches, in alignment with our structuring of cognitive prerequisites and mechanisms for the problem-solving abilities of intelligent biological agents (see \autoref{fig:cog_prereq}). 
We summarise our main results as follows:

\textbf{Most current few-shot problem-solving methods build on transfer learning but they do not leverage planning.}
This result is interesting because any approach with a forward model could straight-forwardly be extended to also consider planning\cite{Botvinick2014}, and planning can be leveraged for few-shot problem-solving\cite{Sharma2020_DADS}. Therefore, current model-based approaches do not exploit their full potential.

\textbf{Current methods do not exploit the full potential of compositional abstract representations.}
All hierarchical reinforcement learning methods inherently perform representational abstraction, e.g., through options or subgoals, but only a few methods consider compositionality. Exceptions include natural language-based representations of actions \cite{Jiang2019_HRL_Language_Abstraction} or symbolic logic-based compositional representations \cite{Oh2017_Zero-shotTaskGeneralizationDeepRL}.
None of these approaches ground these abstractions tightly to sensorimotor experiences.

\textbf{State abstraction is significantly less researched than action abstraction.}
Recent hierarchical actor-critic approaches \cite{Levy2019_Hierarchical,Roeder2020_CHAC} use the same state representation for each layer in the hierarchy, without performing any abstraction. 
There exist a few approaches that perform state abstraction\cite{Vezhnevets2017_Feudal}, and also some that involve compositional state representations\cite{Vezhnevets2020_OPRE}. However, most of these build on hand-crafted abstraction functions that are not learned \cite{Eppe2019_planning_rl}.

\textbf{Curiosity and diversity, as intrinsic motivation methods, are underrepresented.}
When assessing the intrinsic motivation mechanisms of the existing approaches, we distinguish between the three most frequently used kinds of intrinsic motivation: curiosity, diversity, and subgoal discovery. 
\autoref{tab:hrl_properties} indicates that curiosity and diversity are underrepresented intrinsic motivation methods, even though there is significant cognitive evidence that these mechanisms are critical for intelligent behaviour in humans \cite{Oudeyer2018_CompuCuriosityLearning}.

Finally, 
\textbf{there is a lack of hierarchical mental simulation methods.}
While numerous non-hierarchical methods show how mental simulation can improve sample efficiency\cite{Deisenroth2011,Ha2018_WorldModels}, our summary shows that reinforcement learning with hierarchical mental simulation is yet to be explored.
As a result, inter-dependencies between sub-tasks are hard to detect. 

\section{Conclusion}
\label{sec:conclusion}
Our review provides an overview of the cognitive foundations of hierarchical problem-solving and how these are implemented in current hierarchical reinforcement learning architectures.
Herein, we focus on few-shot problem-solving -- the ability to learn to solve causally non-trivial problems with as few trials as possible or, ideally, without any trials at all (zero-shot problem-solving). 
As a benchmark example, we refer to recent results from animal cognition \cite{Gruber2019_CrowMentalRep}, demonstrating how a crow develops mental representations of tools to solve a previously unseen food-access problem at the first trial (see \autoref{fig:crow_probsolving}). 
As our main research question, we ask for the computational prerequisites and mechanisms to enable problem-solving for artificial computational agents that are on the level of intelligent animals. 

We provide two main contributions to address this question. First, we perform a literature survey and a top-down structuring of cognitive abilities, mechanisms, and prerequisites for few-shot problem-solving (see \autoref{fig:cog_prereq}). Second, we perform a comprehensive review of recent hierarchical reinforcement learning architectures to identify whether and how the existing approaches implement the prerequisites, mechanisms, and abilities towards few-shot problem-solving (see \autoref{tab:hrl_properties}).
Herein, we identify five major shortcomings in the state of the art of the computational approaches. 
These shortcomings are mostly based on the lack of hierarchical forward models and compositional abstractions. 

Nonetheless, we were able to identify several methods that address these gaps in the state of the art in isolation. 
Not seeing any major reason why these approaches could not be integrated, we suggest that most of the tools to realise higher levels of intelligent behaviour in artificial agents have already been investigated. 
The key is to combine them effectively. 
This demonstrates significant potential to develop intelligent agents that can learn to solve problems on the level of intelligent biological agents.

\section*{Acknowledgements}
Manfred Eppe, Phuong Nguyen and Stefan Wermter acknowledge funding by the German Research Foundation (DFG) through the projects IDEAS and LeCAREbot. 
Stefan Wermter and Matthias Kerzel acknowledge funding by the DFG through the transregio project TRR 169 Cross-modal Learning (CML).
Christian Gumbsch and Martin Butz acknowledge funding by the DFG project number BU 1335/11-1 in the framework of the SPP program ``The Active Self" (SPP 2134).
Furthermore, Martin Butz acknowledges support from a Feodor-Lynen stipend of the Humboldt Foundation as well as from the DFG, project number 198647426, ``Research Training Group 1808: Ambiguity - Production and Perception'' and from the DFG-funded Cluster of Excellence ``Machine Learning  -- New Perspectives for Science´´, EXC 2064/1, project number 390727645.
The authors thank the International Max Planck Research School for Intelligent Systems (IMPRS-IS) for supporting Christian Gumbsch.

\bibliography{references_final}



\end{document}